\documentclass[10pt,twocolumn,letterpaper]{article}

\usepackage{cvpr}              

\usepackage{amsmath,amsfonts,amssymb,bbm,float}
\usepackage{graphicx}

\usepackage[dvipsnames]{xcolor}
\usepackage{amssymb}
\usepackage{amsmath,amsfonts,amssymb,bbm,float}
\usepackage{amsmath}
\usepackage{cases}
\usepackage{subcaption} 
\usepackage{multirow} 
\usepackage{booktabs}\let\cline\cmidrule

\usepackage{amsmath,amsfonts,amssymb,bbm,float}
\usepackage{graphicx}

\newcommand{\Amat}{{\bf A}}

\newcommand{\Emat}[0]{{{\bf E}}}
\newcommand{\Fmat}[0]{{{\bf F}}}

\newcommand{\Kmat}[0]{{{\bf K}}}

\newcommand{\Qmat}[0]{{{\bf Q}}}

\newcommand{\Vmat}[0]{{{\bf V}}}
\newcommand{\Wmat}[0]{{{\bf W}}}
\newcommand{\Xmat}{{\bf X}}
\newcommand{\Ymat}[0]{{{\bf Y}}}
\newcommand{\Zmat}{{\bf Z}}

\newcommand{\xv}{\boldsymbol{x}}
\newcommand{\yv}{\boldsymbol{y}}

\newcommand{\zv}{\boldsymbol{z}}

\newcommand{\Phimat}{\boldsymbol{\Phi}}
\newcommand{\Psimat}{\boldsymbol{\Psi}}

\usepackage{cases}
\usepackage{amsmath}

\usepackage{amssymb}
\usepackage{url}            
\usepackage{booktabs}       
\usepackage{nicefrac}       
\usepackage{microtype}      
\usepackage{xcolor}         
\usepackage{graphicx}
\usepackage{caption}
\usepackage{tikz}
\usepackage{enumitem}
\usepackage{balance}
\usepackage{multirow}                
\usepackage{multicol}              
\usepackage{float}                   
\usepackage{makecell}                  
\usepackage{threeparttable}

\newcommand{\ts}{^{\top}}

\newcommand{\vc}{{\rm vec}}

\definecolor{cvprblue}{rgb}{0.21,0.49,0.74}
\usepackage[pagebackref,breaklinks,colorlinks,citecolor=cvprblue]{hyperref}

\def\paperID{15809} 
\def\confName{CVPR}
\def\confYear{2024}

\title{Dual-Scale Transformer for Large-Scale Single-Pixel Imaging}

\author{
Gang Qu\textsuperscript{\rm 1,}\footnotemark[1]
\qquad 
Ping Wang\textsuperscript{\rm 1,2,}\thanks{Equal contribution.} 
\qquad 
Xin Yuan\textsuperscript{\rm 1,}\thanks{Corresponding author.}
\\
\textsuperscript{\rm 1}School of Engineering, Westlake University \quad 
\textsuperscript{\rm 2}Zhejiang University
\\
{\tt\small 
\{qugang,wangping,xyuan\}@westlake.edu.cn}
}

\begin{document}
\maketitle
\begin{abstract}
Single-pixel imaging (SPI) is a potential computational imaging technique which produces image by solving an ill-posed reconstruction problem from few measurements captured by a single-pixel detector. 
Deep learning has achieved impressive success on SPI reconstruction. 
However, previous poor reconstruction performance and impractical imaging model limit its real-world applications.
In this paper, we propose a deep unfolding network with hybrid-attention Transformer on Kronecker SPI model, dubbed HATNet, to improve the imaging quality of real SPI cameras.
{Specifically, we unfold the computation graph of the iterative shrinkage-thresholding algorithm (ISTA) into two alternative modules: efficient tensor gradient descent and hybrid-attention multi-scale denoising.
By virtue of Kronecker SPI, the gradient descent module can avoid high computational overheads rooted in previous gradient descent modules based on vectorized SPI.
The denoising module is an encoder-decoder architecture powered by dual-scale spatial attention for high- and low-frequency aggregation and channel attention for global information recalibration.}
Moreover, we build a SPI prototype to verify the effectiveness of the proposed method.
Extensive experiments on synthetic and real data demonstrate that our method achieves the state-of-the-art performance. The source code and pre-trained models are available at \url {https://github.com/Gang-Qu/HATNet-SPI}.

\end{abstract}


\section{Introduction}
\label{sec:intro}

\begin{figure}
  \centering
  \vspace{1mm}
  \includegraphics[width=1\linewidth]{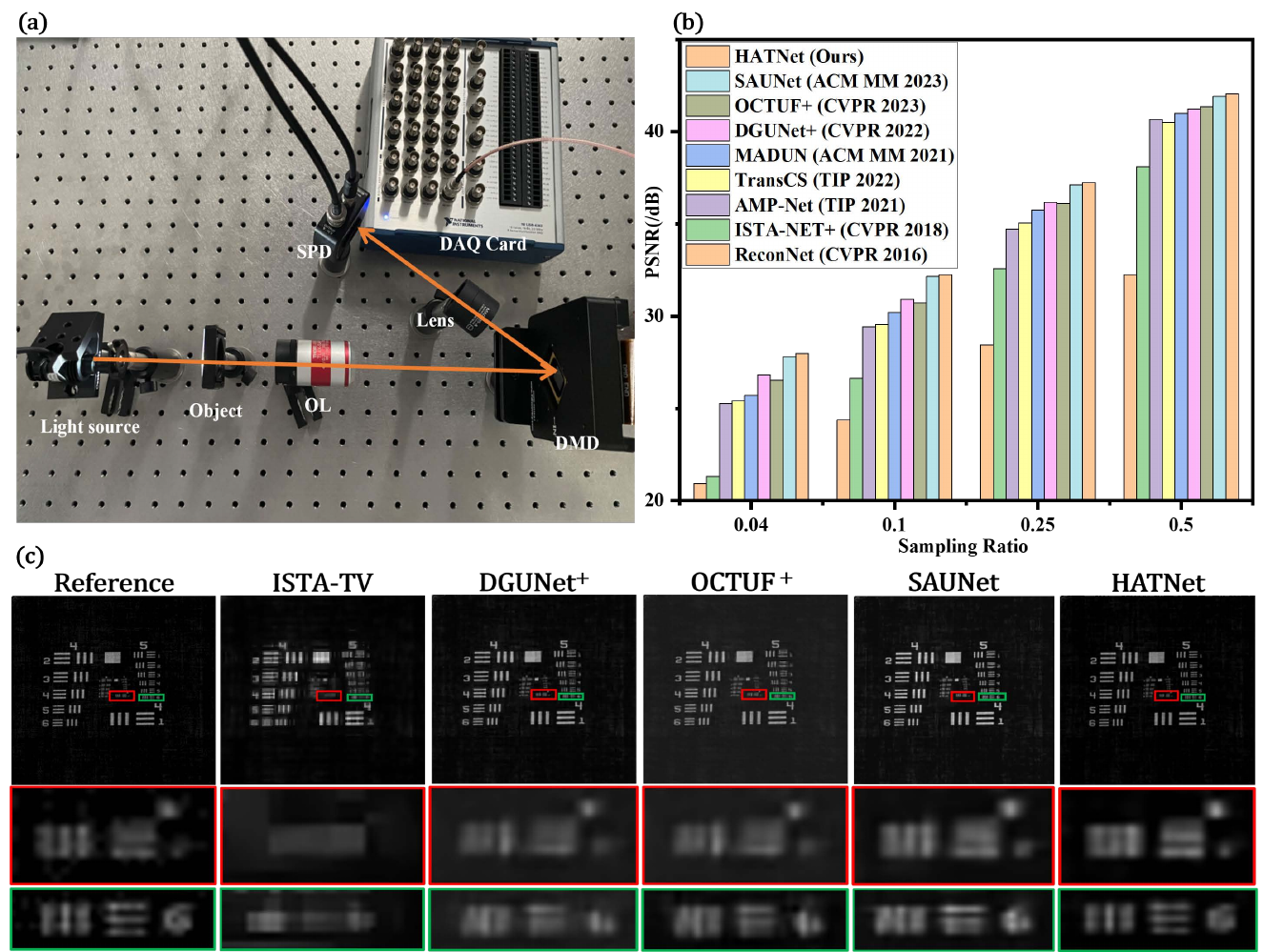}
  \vspace{-4mm}
  \caption{(a) Our built SPI prototype (OL: objective lens, DMD: digital micromirror device, DAQ card: data acquisition card). (b) Performance comparison of different methods on Set11 dataset at different sampling ratios. (c) Real experimental results of different methods at sampling ratio of 25\%.}
\label{fig: setup}
\vspace{-4mm}
\end{figure}

{Conventional imaging technology produces images by exploiting the light reflected or scattered by an object on a two-dimensional (2D) CCD or CMOS detector with millions of pixels.
But in applications, such as the infrared or deep ultraviolet sensing, the availability of pixelated array detectors becomes expensive or impractical.
As an alternative solution, single-pixel imaging (SPI) utilizes just one light-sensitive single-pixel detector (SPD) to record the total intensity of the reflected or scattered light encoded by temporally varying modulation patterns from a spatial light modulator, yielding compressed measurements, and the desired image can be estimated from the captured (compressed) measurements via iterative optimization algorithms or a deep learning model.
SPI camera offers advantages over conventional cameras, such as improved detection efficiency, lower dark counts, and faster timing response.
Such advantages can have significance in scenarios where the detected intensities are very weak due to scattering or absorption losses.
Moreover, SPI camera is capable of sensing compressively during data acquisition, thereby reducing the data storage and communication bandwidth requirements.
During the last decade, SPI has been widely used in} 
3D imaging~\cite{1}, hyperspectral imaging~\cite{xu2024compressive}, X-ray diffraction tomography~\cite{greenberg2014compressive}, magnetic resonance imaging~\cite{haldar2010compressed}, ophthalmic imaging~\cite{lochocki2016single} and imaging in non-visible wavebands~\cite{8} or through scattering media~\cite{9}.

{SPI is driven from the compressive sensing (CS)~\cite{candes2006robust,donoho2006compressed,Yuan18OE, slope,yuan2020image,wang2023deep,wang2023full} theory.
In CS paradigm, a 1D signal $\xv \!\in\! \mathbbm{R}^{N}$ is compressively sampled into few measurements $\yv \!\in\! \mathbbm{R}^{M}$ at a sub-Nyquist sampling ratio $\frac{M}{N}$ ($M \!\ll\! N $) via a linear system:
\begin{equation}
\label{eq:forward}
\setlength{\abovedisplayskip}{0.25cm}
\setlength{\belowdisplayskip}{0.25cm}
\yv = \Amat \xv,
\end{equation}
where $\Amat \!\in\! \mathbbm{R}^{M\times N}$ is the measurement matrix. Due to the ill-posed nature of the inverse process of \cref{eq:forward}, an estimate $\hat{\xv}$ of $\xv$ could be reconstructed from $\yv$ by solving the following optimization problem:
\begin{equation}
\label{eq:backward}
\setlength{\abovedisplayskip}{0.25cm}
\setlength{\belowdisplayskip}{0.25cm}
\textstyle \hat{\xv} = \mathop{\arg\min}\limits_{\xv}\frac{1}{2}\Vert \yv - \Amat\xv \Vert^2_2+\lambda \mathcal{R}(\xv), 
\end{equation}
where $\frac{1}{2}\Vert \yv-\Amat\xv \Vert^2_2$ is a data fidelity term, and $\lambda\mathcal{R}(\xv)$ is a regularization term.
Over the past years, various optimization algorithms have been developed to solve it, among which iterative shrinkage-thresholding algorithm (ISTA)~\cite{daubechies2004iterative} is the most widely used one. ISTA is composed of two alternative operations:
\begin{numcases}{}
\setlength{\abovedisplayskip}{0.17cm}
\setlength{\belowdisplayskip}{0.17cm}
\label{eq:ista1}
{\zv}_{k}=\mathbf{\xv}_{k-1} + \rho{{\Amat\ts} \left({\yv} - \Amat{\xv}_{k-1} \right)}, \\
\label{eq:ista2}
{\xv}_{k}=\underset{{\xv}}{\arg\min} \frac{1}{2\sigma^2}||{\zv}_{k} - {\xv}||_{2}^{2}+ \mathcal{R}({\xv}),
\end{numcases}
where $\rho = (1+\eta)^{-1}$ and $\sigma = {\sqrt {\lambda \mathord{\left/ {\vphantom {\lambda \eta}} \right. \kern-\nulldelimiterspace} \eta}}$ both with a penalty parameter $\eta$.
\cref{eq:ista1} is a gradient descent process and  Eq.~\eqref{eq:ista2} is equivalent to denoising image $\zv_k$ with the regularization term $\mathcal{R}({\xv})$.
Regarding $\mathcal{R}({\xv})$, various hand-crafted image priors have been proposed to regularize the solution in the desired signal space, such as sparsity~\cite{potter2010sparsity, sparsity-dl}, total variation~\cite{GAPTV}, low rank~\cite{rank}, and non-local self-similarity~\cite{zha2020group}. Unfortunately, hundreds and thousands of iterations lead to just passable results, thereby making it   impractical for real-time and high-fidelity scenarios.
Over the past few years, deep neural networks have recently gained considerable popularity in solving the inverse problem of CS, from early black-box networks~\cite{kulkarni2016reconnet,lyu2017deep,shi2019scalable,yao2019dr2} to recent deep unfolding networks (DUNs)~\cite{zhang2018ista,zhang2020optimization,zhang2020amp,shen2022transcs,song2021memory,mou2022deep,meng2023deep}.
Early black-box networks usually learn a non-linear mapping from sampled measurements to the final reconstructed result in an end-to-end manner, with limited performance and without interpretability.
By combining an optimization algorithm with a deep denoising network, DUNs enjoy the interpretability of optimization algorithms and the powerful modeling ability of deep neural networks, leading to the-state-of-art (SOTA) performance.
These deep models, both black-box networks and DUNs, are developed as CS solvers.}

{However, there is a significant mismatch between real SPI system and CS-oriented solvers. Regular CS model in Eqs.~\eqref{eq:forward} to \eqref{eq:ista2} is established for vectored 1D signal but SPI cameras take aim at 2D image. 
Before this work, 2D image to be detected is considered as the vectorized signal when a CS-oriented solver is employed, leading a huge measurement matrix $\Amat$ and thus extremely high computational cost in~\cref{eq:ista1}.  
For example, a $512\times512$ image needs a measurement matrix $\Amat$ with $262,144$ columns, which is terribly large.
To address this problem, most previous DUNs~\cite{zhang2018ista,zhang2020optimization,zhang2020amp,shen2022transcs,song2021memory,mou2022deep} divide the whole image into several small patches to process, thus also called block-CS-oriented solvers.
However, block-based sampling is impractical for mainstream SPI cameras. This is why these DUNs are outstanding on simulation metrics but are rarely deployed on real-world SPI cameras.
Recently, the first practical DUN~\cite{wang2023saunet} has been proposed to enable full image CS reconstruction but also overlooks physical constraints in imaging.}

{To bridge the gap between SPI and DUNs, we propose a deep unfolding network with hybrid-attention Transformer, dubbed HATNet, on Kronecker SPI~\cite{duarte2011kronecker} by unrolling the computation graph of ISTA into two alternative modules: efficient gradient descent and HAT-based denoising.   
The main contributions of this work are summarized as follows.}

\begin{itemize}
    \item [1)] To exploit global interactions of images and satisfy physical constraints of real SPI cameras, we introduce a DUN of tensor ISTA, composed of tensor gradient descent module and deep denoising module, to enable full-size sampling and reconstruction for SPI. By avoiding vectorized huge measurement matrix, the forward model of SPI and the gradient descent of DUN are significantly accelerated. 
    \item [2)] We propose a DUN with hybrid-attention Transformer, dubbed HATNet, powered by spatial-wise dual-scale self-attention and channel-wise self-attention. HAT is capable of aggregating high- and low-frequency information and recalibrating channel-wise global information.
    \item [3)] We use HAT under an encoder-decoder architecture as the deep denoiser of deep unfolding and it achieve SOTA performance on synthetic data as reported in \cref{fig: setup} (b). Moreover, we also verify the effectiveness of proposed method on real data as demonstrated in \cref{fig: setup} (c), which is captured by our SPI prototype in \cref{fig: setup} (a).
    To our best knowledge, we are the first to develop a SOTA deep model to improve practical SPI, particularly for large scale.   
\end{itemize}

%

\section{Related Work}
\label{sec:formatting}

\subsection{Compressive Sensing Reconstruction}
CS reconstruction methods could be classified into two categories: optimization-based methods~\cite{figueiredo2007gradient,4587391,he2009exploiting,blumensath2009iterative,beck2009fast,kim2010compressed,yang2011alternating,dong2014compressive,zhang2014group,Metzler2016FromDT} and learning-based methods~\cite{kulkarni2016reconnet,metzler2017learned,zhang2018ista,yang2018admm,shi2019image,shi2019scalable,yao2019dr2,zhang2020optimization,zhang2020amp,shen2022transcs,song2021memory,mou2022deep,song2023optimization}. {Optimization-based methods mainly employ an iterative optimization algorithm along with hand-crafted image priors to increasingly retrieve the visual information from the sub-sampled measurement}.
Various iterative optimization algorithms have been proposed, including iterative shrinkage-thresholding algorithm (ISTA)~\cite{daubechies2004iterative}, 
approximate message passing (AMP) algorithm~\cite{zhu2020deformable}, alternating direction method of multiplies (ADMM)~\cite{goldstein2014fast}, generalized alternating projection (GAP)~\cite{GAPTV} method, least absolute shrinkage and selection operator (LASSO)~\cite{tibshirani1996regression}.
TVAL3 \cite{li2013efficient} utilizes the augmented Lagrangian method with total variation (TV) prior to remove the noise and restore the details.
However, optimization-based methods need hundreds and thousands of iterations and usually have a long processing time and limited reconstruction quality. 
In recent years, deep neural networks have been developed as powerful CS solvers and have achieved impressive success. Early networks~\cite{kulkarni2016reconnet,metzler2017learned} usually learn a black-box mapping from the compressed measurements to the restored image. Most recently, deep unfolding networks~\cite{zhang2018ista,zhang2020optimization,zhang2020amp,shen2022transcs,mou2022deep,meng2023deep} use a deep denoising network to replace the proximal mapping and maintain the gradient descent of a conventional optimization algorithms, which achieve SOTA performance after few iterations.
With good performance and interpretability,
DUNs have become the mainstream choice for CS reconstruction.
Such regime was originally applied in plug-and-play (PnP) methods, where pre-trained denoiser is employed to implicitly express the regularization term as a denoising problem~\cite{meinhardt2017learning}.
Different iterative optimization algorithms foster kinds of DUNs, such as ADMM-Net~\cite{zhang2020amp}, ISTA-Net~\cite{zhang2018ista}, AMP-Net~\cite{zhang2020amp}.
However, most of previous DUNs are troubled with the information loss rooted in the frequent {signal-to-feature transformation}. This problem indicates that
early-stage high-level features cannot be efficiently used for later-stage feature refinement.
Latest DUNs~\cite{song2021memory,song2023optimization, song2023dynamic, wang2023saunet} try to solve this problem by heuristic cross-stage information fusion designs.
In addition, most of previous DUNs are developed under the block-based sampling assumption, which is impractical for real SPI.

\subsection{Vision Transformer}
Motivated by the power of Transformer \cite{vaswani2017attention} in natural language processing (NLP), ViT~\cite{arnab2021vivit} first extends Transformer into vision tasks by conducting self-attention (SA) mechanism on non-overlapping patches. Swin Transformer~\cite{liu2021swin} proposes a pioneering SA within shifted windows under a hierarchical architecture to achieve significant improvement over convolutional neural networks (CNNs) on kinds of vision tasks. Due to the remarkable performance of SA, researchers are extending Transformer into low-level vision tasks~\cite{fan2021multiscale,yao2023dual, pan2022fast,chen2021pre,wang2022uformer, zamir2022restormer}. PIT first introduces Transformer to image restoration and showcases its performance on several image restoration tasks \cite{chen2021pre}. Uformer \cite{wang2022uformer} combines Transformer and U-Net to build multi-scale Transformer to further improve the performance.
TransGAN, a mixture of generative adversarial network (GAN) and Transformer, is proposed in \cite{jiang2021transgan} for image generation.
Restormer~\cite{zamir2022restormer} operates self-attention along channel dimension for high-resolution image restoration.
These Transformer-based methods remarkably outperform CNN-based methods and also reveal that the attention mechanism on both spatial and channel dimensions are significant for most vision tasks.

\section{Proposed Method}

\begin{figure*}[t]
\centering
\includegraphics[width=1\linewidth]{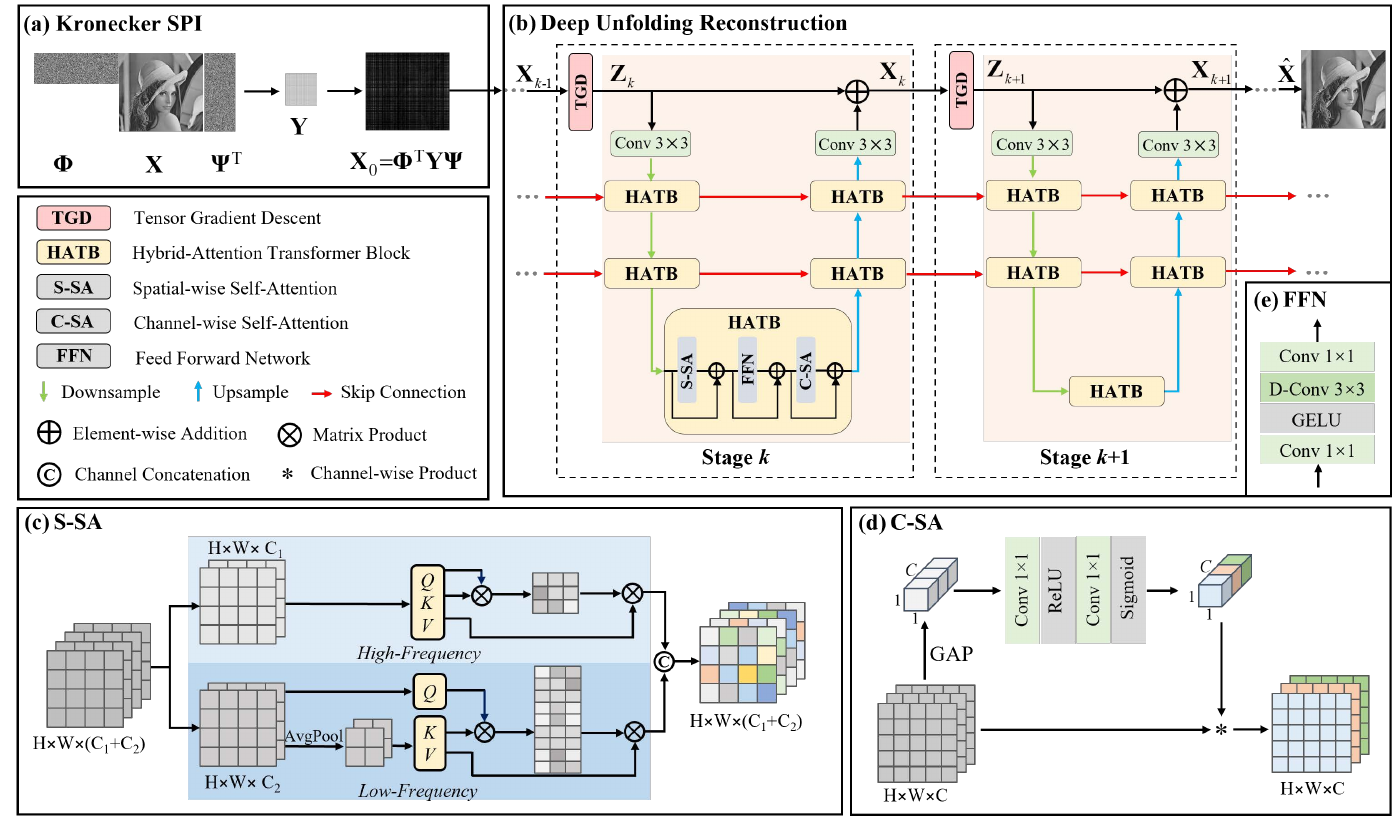}
\caption{Illustration of the proposed method. (a) demonstrates the Kronecker SPI model. As shown in (b), our DUN aims to reconstruct a high-fidelity image $\hat \Xmat$ from the initialization input $\Xmat_0$, which is composed of multiple stages with skip connections and each stage involves a tensor gradient descent (TGD) operator in \cref{eq: z11} and a U-shaped deep denoiser as Eq.~\eqref{eq: x11}.
The deep denoiser is powered by the proposed HATB, each of which consists of residual dual-scale spatial-wise self-attention (S-SA), feed-forward network (FFN), and channel-wise self-attention (C-SA). The structure of S-SA and C-SA are shown in (c) and (d), respectively.}
\label{fig: arch}
\vspace{-2mm}
\end{figure*}


\subsection{Tensor ISTA Unfolding Framework}



{In SPI paradigm, assume $\Xmat \!\in\! \mathbbm{R}^{{\sqrt N} \times {\sqrt N}}$ is a 2D image, and its 2D compressed measurements $\Ymat \!\in\! \mathbbm{R}^{{\sqrt M}\times {\sqrt M}}$ can be obtained by a linear measurement system:
\begin{equation}
\label{eq:2}
\setlength{\abovedisplayskip}{0.25cm}
\setlength{\belowdisplayskip}{0.25cm}
\Ymat = \Phimat \Xmat \Psimat^\top.
\end{equation}
where $\Phimat \!\in\! \mathbbm{R}^{{\sqrt M}\times {\sqrt N}}, \Psimat \!\in\! \mathbbm{R}^{{\sqrt M}\times {\sqrt N}}$ are two independent measurement matrices, simultaneously compressing image along horizontal and vertical dimensions.
In real imaging systems, such dual modulation is impossible to implement.
\cref{eq:2} is equivalent to the vectorized CS form in \cref{eq:forward} through the Kronecker product~\cite{duarte2011kronecker}:
\begin{equation}
\begin{aligned}
\label{eq:3}
\yv = \Amat \xv, \quad s.t. \quad 
\left\{ {\begin{array}{*{20}{c}}
\xv = \vc \left(\Xmat\right),\\
\yv = \vc \left(\Ymat\right),\\
\Amat = \Psimat \otimes \Phimat,
\end{array}} \right.
\end{aligned}
\end{equation}
where vec(·) denotes the vectorization operation and $\otimes$ represents the Kronecker product. The ill-posed inverse process of \cref{eq:2} can be conducted by solving the following optimization problem:}
\begin{align} \label{eq: dec}
 {\hat \Xmat} =\underset{\Xmat}{\arg\min}~\frac{1}{2} \left \| {\Ymat} - \Phimat \Xmat {\Psimat^{\top}} \right \|_{F}^{2}+ \lambda \mathcal{R}({\Xmat}),
\end{align}
where ${\left\|  \cdot  \right\|_{F}}$ denotes the Frobenius norm.
The above optimization problem can be solved by the tensor version of ISTA~\cite{wang2023saunet}, namely 
\begin{numcases}{}
{\label{eq: z}
\Zmat_{k}  =  \Xmat_{k-1} + \rho  {\Phimat^\top(\Ymat - \Phimat \Xmat_{k-1} \Psimat^\top)\Psimat},} \\
{\label{eq: x}
{\Xmat}_{k} = \underset{\Xmat}{\arg\min}~\frac{1}{2{\sigma^2}} || \Zmat_{k} - \Xmat  ||_{F}^{2} + \mathcal{R}(\Xmat),}
\end{numcases}
\cref{eq: z} is a tensor gradient descent (TGD) with a step size $\rho$. Eq.~\eqref{eq: x} is a proximal mapping, usually can be seen as a denoising problem~\cite{chan2016plug} with the noise level $\sigma$ from the perspective of Bayesian probability.
By alternatively repeating the above two steps enough times, a decent estimate ${\hat \Xmat}$ would well approach to the ground truth ${\Xmat}$.
In this manner, image reconstruction task in \cref{eq: dec} is converted to a multi-stage image denoising task, which has extensively studied in low-level vision~\cite{liu2021swin,Liang2021ICCV}.
Recently, learning-based denosiers have shown great performance gains over conventional optimization-based denosiers.

By unfolding the computation graph of tensor ISTA into deep neural network, we propose a deep unfolding network with hybrid-attention
Transformer (HATNet), which can be formulated as
\begin{numcases}{}
\label{eq: z11}
\Zmat_{k}  =  \Xmat_{k-1} + \rho_{k-1} {\Phimat^\top(\Ymat - \Phimat \Xmat_{k-1} \Psimat^\top)\Psimat}, \\
\label{eq: x11}
{\Xmat}_{k} = {\mathcal D}_{(\theta, k)}(\Zmat_{k}),
\end{numcases}
where $\rho_{k-1}$ is a learnable step size controlling the intensity of $k$-th gradient descent and ${\mathcal D}_{(\theta, k)}$ is a stage-specific deep denoiser with learnable parameters $\theta$.
The initialization input is $\Xmat_{0} = \Phimat^\top\Ymat \Psimat$.
Regarding the design of ${\mathcal D}_{(\theta, k)}$, we propose a hybrid-attention Transformer (HAT) as building block, where spatial-wise dual-scale attention for long-range high- and low-frequency aggregation and channel-wise attention for global information recalibration are established, which will be introduced in detail in \cref{sec: denoiser}.

\subsection{Deep Denoiser with HAT}
\label{sec: denoiser}
In this sub-section, we give the details of deep denoiser ${\mathcal D}_{(\theta, k)}$ used in Eq.~\eqref{eq: x11}.
Different-stage denoisers have the same network structure with independent learnable parameters and thus we introduce just one of it.


\noindent{\bf Overall Architecture.}
As shown in \cref{fig: arch} (b), $k$-th stage denoiser is a symmetric encoder-decoder architecture built by multiple hybrid-attention Transformer blocks (HATBs) to generate the residual image with degraded input. Each HATB is powered by a residual spatial-wise self-attention (S-SA), feed-forward network (FFN), and channel-wise self-attention (C-SA).
As illustrated in \cref{fig: arch} (e), FFN is composed of two $1\times1$ convolutions which increases or decreases the channel dimensions, one $3\times3$ depth-wise convolution (D-Conv), and one non-linear activation GELU between them.
The downsampling layer uses a $2\times2$ convolution with a stride of $2$. The upsampling layer uses a point-wise convolution ($1\times1$ Conv) to double the channel dimensions and then is followed by a pixel shuffle operation.
At each stage, the encoder features are concatenated with the decoder features via skip connections and then a point-wise convolution is used to reduce channel dimensions by half for efficient feature fusion and refinement.
At two adjacent stages, previous-stage decoder features are also fused with current-stage encoder features to avoid the potential information loss caused by the signal-feature transformation of deep unfolding.
Such dense skip connections between both intra-stage and inter-stage HATBs enhance the performance of proposed method clearly as demonstrated in the ablation experiments shown in \cref{tab: ablation}.
As two core components of the proposed HAT, S-SA and C-SA can realize spatial high- and low-frequency aggregation and channel-wise recalibration respectively.
Next, we will describe the details.

\noindent{\bf Spatial-wise Self-Attention (S-SA).} 
S-SA conducts multi-head self-attention mechanism on dual scales within shifted windows.
Specifically, given the input feature $\Fmat \!\in\! \mathbbm{R}^ {H \times W \times C}$, S-SA generates two groups of {\em query}, {\em key}, and {\em value} through the following linear projection:
\begin{numcases} {}
\label{eq: Q1}
\Qmat^{h} =  \Fmat\Wmat^{h}_q, \qquad 
\Qmat^{l} = \Fmat \Wmat^{l}_q,\\
\Kmat^{h} =   \Fmat\Wmat^{h}_k, \qquad
\Kmat^{l} = {\tt AvgPool}(\Fmat) \Wmat^{l}_k, \\ 
\Vmat^{h} =   \Fmat\Wmat^{h}_v, \qquad
\Vmat^{l} = {\tt AvgPool}(\Fmat) \Wmat^{l}_v,
\end{numcases}
where $\Wmat^{h}_q,\Wmat^{h}_k,\Wmat^{h}_v \!\in\! \mathbbm{R}^{C \times C_1}$ and $\Wmat^{l}_q,\Wmat^{l}_k,\Wmat^{l}_v \!\in\! \mathbbm{R}^{C \times C_2}$ are learnable projection matrices with biases omitted, and ${\tt AvgPool}$ represents an average pooling operator with the window resolution $p$.
The resulting $\Qmat^{h},\Kmat^{h},\Vmat^{h} \!\in\! \mathbbm{R}^ {H \times W \times C_{1}}$ belong to the high-frequency group, similar to that of regular Transformer~\cite{liu2021swin}. The resulting $\Qmat^{l}\!\in\! \mathbbm{R}^ {H \times W \times C_{2}}$ and $\Kmat^{l},\Vmat^{l} \!\in\! \mathbbm{R}^ {h \times w \times C_{2}}$ belong to low-frequency group, similar to that of PVT~\cite{wang2021pyramid}, where $h \!=\! \frac{H}{\sqrt p}$ and $w \!=\! \frac{W}{\sqrt p }$.
The total channel number of high-frequency and low-frequency branches is the same with that of the input feature, namely $C_1 \!+\! C_2 \!=\! C$.
Then, $\Qmat^{h},\Kmat^{h},\Vmat^{h}$ and $\Qmat^{l},\Kmat^{l},\Vmat^{l}$ are partitioned into non-overlapping windows and then flatted into token sequences.
For high-frequency {\em query}, {\em key}, and {\em value}, the window resolution is $N$ and the reshaped results can be represented as ${\bar\Qmat}^{h},{\bar\Kmat}^{h},{\bar\Vmat}^{h} \!\in\! \mathbbm{R}^ {\frac{HW}{N} \times N \times C_{1}}$.
For low-frequency {\em query}, {\em key}, and {\em value}, the window resolution is $pN$ for $\Qmat^{l}$ and $N$ for $\Kmat^{l},\Vmat^{l}$, and the reshaped results can be represented as ${\bar\Qmat}^{l} \!\in\! \mathbbm{R}^ {\frac{hw}{N} \times pN \times C_{1}}$ and ${\bar\Kmat}^{l},{\bar\Vmat}^{l} \!\in\! \mathbbm{R}^ {\frac{hw}{N} \times N \times C_{1}}$. 
Next, they are split into $m$ heads, namely $\{{\bar\Qmat}^{h}_i\}_{i=1}^{m}$, $\{{\bar\Kmat}^{h}_i\}_{i=1}^{m}$, $\{{\bar\Vmat}^{h}_i\}_{i=1}^{m}$, $\{{\bar\Qmat}^{l}_i\}_{i=1}^{m}$, $\{{\bar\Kmat}^{l}_i\}_{i=1}^{m}$, $\{{\bar\Vmat}^{l}_i\}_{i=1}^{m}$.
The channel dimension of each head is $d\!=\!\frac{C_1}{m}$ for high-frequency group and $d\!=\!\frac{C_2}{m}$ for low-frequency group.
The illustration of \cref{fig: arch} (c) is the case with $m\!=1\!$.
For $i$-th head, high-frequency output ${{\bar\Emat}^{h}_i}$ and low-frequency output ${{\bar\Emat}^{l}_i}$ are calculated by
\begin{equation}
\begin{aligned}
\label{eq: SA}
{{\bar\Emat}^{h}_i} &= {\tt softmax}(\frac{{{\bar\Qmat}^{h}_i} {{\bar\Kmat}^{h \top}_i}}{\sqrt{d}} ) {{\bar\Vmat}^{h}_i}, \\
{{\bar\Emat}^{l}_i} &= {\tt softmax}(\frac{{{\bar\Qmat}^{l}_i} {{\bar\Kmat}^{l \top}_i}}{\sqrt{d}} ) {{\bar\Vmat}^{l}_i}.
\end{aligned}
\end{equation}
As a result, high-frequency feature ${\Emat}^{h} \!\in\! \mathbbm{R}^ {H \times W \times C_{1}}$ and low-frequency feature ${\Emat}^{l} \!\in\! \mathbbm{R}^ {H \times W \times C_{2}}$ can be got by reshaping and concatenating $\{{{\bar\Emat}^{h}_i}\}_{i=1}^{m}$ and $\{{{\bar\Emat}^{h}_i}\}_{i=1}^{m}$ separately.
The final output is obtained by fusing ${\Emat}^{h} \!\in\! \mathbbm{R}^ {H \times W \times C_{1}}$ and ${\Emat}^{l} \!\in\! \mathbbm{R}^ {H \times W \times C_{2}}$. This process is formulated as
\begin{equation}
\begin{aligned}
{\rm S\!\!-\!\!SA}(\Fmat) = {\tt Concat}( {\Emat}^{h} {\Wmat_{h}}, {\Emat}^{l} {\Wmat_{l}}),
\end{aligned}
\end{equation}
where $\Wmat^{h} \!\in\! \mathbbm{R}^{C_{1} \times C_{1}},\Wmat^{l}\!\in\! \mathbbm{R}^{C_{2} \times C_{2}}$ are two learnable projection matrices and ${\tt Concat}$ denotes the channel concatenation.

As illustrated in \cref{fig: arch} (c), the high-frequency attention performs regular attention within $N$-pixel windows, and the low-frequency attention performs cross-scale attention between  {\em query} and average-pooled {\em key}, {\em value} within $pN$-pixel windows.
As average pooling can act as a low-pass filter~\cite{pan2022fast}, such dual-scale attention has two sizes of receptive fields on the input and the average-pooled input, enabling high- and low-frequency aggregation.

\noindent{\bf Channel-wise Self-Attention (C-SA).} 
Since that S-SA focuses on capturing spatial information within local windows, we incorporate a channel-wise self-attention (C-SA) to capture channel-wise global information~\cite{hu2018squeeze}.
As illustrated in \cref{fig: arch} (d), C-SA squeezes the spatial information into channels first and then a multilayer perceptron applies to it to calculate the channel attention, which will be used to weight the feature map. Given an input $\Fmat \!\in\! \mathbbm{R}^ {H \times W \times C}$ , the output of C-SA is formulated as
\begin{equation}
\begin{aligned}
{\rm C\!\!-\!\!SA}(\Fmat) = {\Fmat} * {\tt Sigmoid}({\tt ReLU}({\tt GAP}(\Fmat)\Wmat_{1})\Wmat_{2}),
\end{aligned}
\end{equation}
where $*$ denotes channel-wise multiplication, $\Wmat_{1} \!\in\! \mathbbm{R}^ {C \times \frac{C}{\beta}}$, $\Wmat_{2} \!\in\! \mathbbm{R}^ {\frac{C}{\beta} \times C}$ are two fully-connected layers with a non-linear activation ${\tt ReLU}$ inside, ${\tt GAP}$ indicates the global average pooling operation, and ${\tt Sigmoid}$ limits the channel attention map in $(0,1)$. $\beta$ is a channel shrinking factor.
C-SA plays two important roles, that is, global information aggregation and channel-wise recalibration.


\section{Experiment}


\begin{table*}[!htbp]
\centering
\caption{Average PSNR/SSIM of different methods on Set11 datasets with different SRs. The best and second best results are highlighted in {\bf bold} and \underline{underlined}, respectively.}
\label{tab: result}
\renewcommand\tabcolsep{6pt}
\scalebox{0.92}{
\begin{tabular}{c|c|cccc}
\toprule[1pt]
\multicolumn{1}{c|}{\multirow{2}{*}{Dataset}}  &
\multicolumn{1}{c|}{\multirow{2}{*}{Method}}   & 
\multicolumn{4}{c}{Sampling Ratio (SR)} 
\\ \cline{3-6}  
\multicolumn{1}{c|}{} &  
\multicolumn{1}{c|}{}  &

\multicolumn{1}{c}{4\%}  & 
\multicolumn{1}{c}{10\%}  & 
\multicolumn{1}{c}{25\%}  & 
\multicolumn{1}{c}{50\%} \\
\toprule[1pt]
\multicolumn{1}{c|}{}  &
\multicolumn{1}{l|}{ReconNet ~\cite{kulkarni2016reconnet}}   & 

\multicolumn{1}{c}{20.93/0.5897} & 
\multicolumn{1}{c}{24.38/0.7301} & 
\multicolumn{1}{c}{28.44/0.8531} & 
\multicolumn{1}{c}{32.25/0.9177} \\
\multicolumn{1}{c|}{}  &
\multicolumn{1}{l|}{ISTA-Net$^+$ 
~\cite{zhang2018ista}}      & 

\multicolumn{1}{c}{21.32/0.6037} & 
\multicolumn{1}{c}{26.64/0.8087} & 
\multicolumn{1}{c}{32.59/0.9254} & 
\multicolumn{1}{c}{38.11/0.9707} \\
\multicolumn{1}{c|}{}  &
\multicolumn{1}{l|}{CSNet$^+$ 
~\cite{shi2019image}}      & 

\multicolumn{1}{c}{24.83/0.7480} & 
\multicolumn{1}{c}{28.34/0.8580} & 
\multicolumn{1}{c}{33.34/0.9387} & 
\multicolumn{1}{c}{38.47/0.9796} \\
\multicolumn{1}{c|}{}  &
\multicolumn{1}{l|}{SCSNet~\cite{shi2019scalable}}      & 

\multicolumn{1}{c}{24.29/0.7589} &
\multicolumn{1}{c}{28.52/0.8616} & 
\multicolumn{1}{c}{33.43/0.9373} &
\multicolumn{1}{c}{39.01/0.9769} \\
\multicolumn{1}{c|}{}  &
\multicolumn{1}{l|}{OPINE-Net$^+$~\cite{zhang2020optimization}} &

\multicolumn{1}{c}{25.69/0.7920} &
\multicolumn{1}{c}{29.81/0.8884} & 
\multicolumn{1}{c}{34.86/0.9509} & 
\multicolumn{1}{c}{40.17/0.9797} \\
\multicolumn{1}{c|}{Set11}  &
\multicolumn{1}{l|}{AMP-Net~\cite{zhang2020amp}} &

\multicolumn{1}{c}{25.27/0.7821} &
\multicolumn{1}{c}{29.43/0.8880} & 
\multicolumn{1}{c}{34.71/0.9532} & 
\multicolumn{1}{c}{40.66/0.9827} \\
\multicolumn{1}{c|}{}  &
\multicolumn{1}{l|}{TransCS~\cite{shen2022transcs}}   & 

\multicolumn{1}{c}{25.41/0.7883	} & 
\multicolumn{1}{c}{29.54/0.8877} & 
\multicolumn{1}{c}{35.06/0.9548} & 
\multicolumn{1}{c}{40.49/0.9815} \\
\multicolumn{1}{c|}{}  &
\multicolumn{1}{l|}{MADUN~\cite{song2021memory}} &

\multicolumn{1}{c}{25.71/0.8042} &
\multicolumn{1}{c}{30.20/0.9016} & 
\multicolumn{1}{c}{35.76/0.9601} & 
\multicolumn{1}{c}{41.00/\underline{0.9837}} \\
\multicolumn{1}{c|}{}  &
\multicolumn{1}{l|}{{DGUNet$^+$~}\cite{mou2022deep}} &
\multicolumn{1}{c}{{26.82/0.8230}} &
\multicolumn{1}{c}{{30.93/0.9088}} & 
\multicolumn{1}{c}{{36.18/0.9616}} & 
\multicolumn{1}{c}{41.24/\underline{0.9837}} \\
\multicolumn{1}{c|}{}  &
\multicolumn{1}{l|}{OCTUF$^+$~\cite{song2023optimization}} &
\multicolumn{1}{c}{26.54/0.8150} &
\multicolumn{1}{c}{30.73/0.9036} & 
\multicolumn{1}{c}{36.10/0.9607} & 
\multicolumn{1}{c}{41.35/\bf{0.9838}} \\

\multicolumn{1}{c|}{}  &
\multicolumn{1}{l|}{SAUNet~\cite{wang2023saunet}} &
\multicolumn{1}{c}{\underline{27.80}/\underline{0.8353}} &
\multicolumn{1}{c}{\underline{32.15}/\underline{0.9147}} & 
\multicolumn{1}{c}{\underline{37.11}/\underline{0.9628}} & 
\multicolumn{1}{c}{\underline{41.91}/\bf{0.9838}} \\

\multicolumn{1}{c|}{}  &
\multicolumn{1}{l|}{{\bf HATNet (ours)}}      & 
\multicolumn{1}{c}{\bf{27.98}/\bf{0.8382}} & 
\multicolumn{1}{c}{\bf{32.26}/\bf{0.9182}} & 
\multicolumn{1}{c}{\bf{37.24}/\bf{0.9634}} &
\multicolumn{1}{c}{\bf{42.05}/\bf{0.9838}} \\

\toprule[1pt]
\end{tabular}
}
\vspace{-3mm}
\end{table*}

\subsection{Implementation Details}
\begin{figure*}[th]
  \centering
  \includegraphics[width=0.95\linewidth]{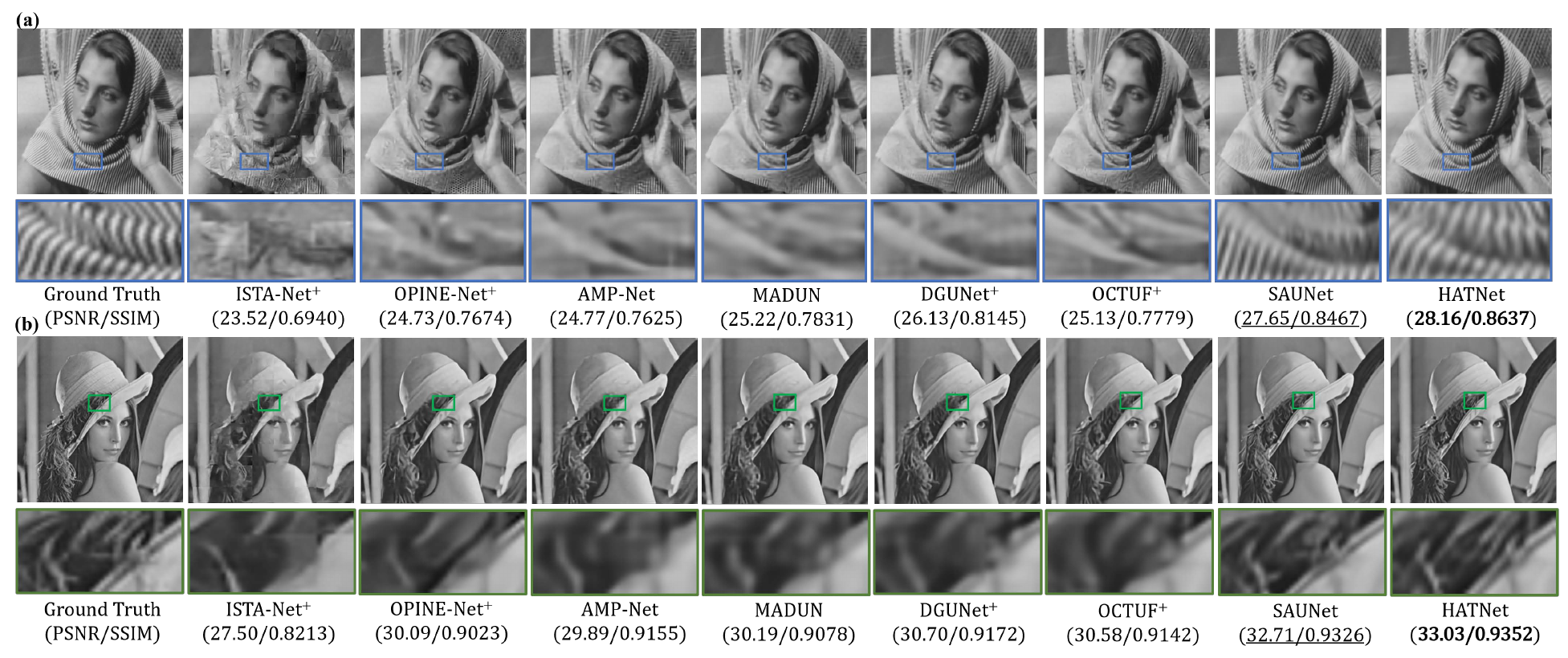}
  \vspace{-3mm}
  \caption{Visualization of different methods on (a) \texttt{Barbara} and (b) \texttt{Lena} at SR = $10\%$.}
\label{fig: SR}
\end{figure*}

In the proposed HATNet, each-stage denoiser is a three-level symmetric U-shaped structure, powered by proposed S-SA and C-SA.
From level-1 to level-3, the number of HATB are $[1,1,1]$ and the dimensions of head is $16$.
Toward S-SA, the size of shifted windows is $4\!\times\!16$ or $16\!\times\!4$ and the kernel size of average pooling operator is $2\!\times\!2$, namely $N\!=\!64$ and $p\!=\!4$.
Toward C-SA, the channel shrinking factor is $\beta\!=\!16$.
Following previous works~\cite{shi2019image,shi2019scalable,song2021memory,zhang2020amp,shen2022transcs,mou2022deep,song2023optimization,wang2023saunet}, we adopt 400 images from BSD500~\cite{5557884} as the training dataset. Data augmentation operations, including random horizontal flipping, random scaling, and random cropping, are performed to generate 20,000 images as the training dataset.
Proposed method is implemented by PyTorch on NVIDIA A100 GPUs. 
All models are trained through 100 epochs with learning rate ${1\times10^{-3}}$ and then fine-tuned through 20 epochs with learning rate ${1\times10^{-4}}$ using Adam optimizer (${\beta_1} = 0.9$, ${\beta_2} = 0.999$).
Similar to previous works~\cite{mou2022deep,song2023optimization,wang2023saunet}, two measurement matrices of Kronecker SPI are set to be learnable for fair comparison on simulation.
In real SPI, they are set to be cake-cutting Hadamard matrices~\cite{cch, hadamardorder, 2022demosaicing}.
For testing on synthetic data, we evaluate the proposed method with different sampling ratios (SRs) $\{4\%,10\%,25\%,50\% \}$ on a commonly-used Set11 dataset.
For testing on real data, we build a SPI prototype to verify the effectiveness of our method.
Peak Signal to Noise Ratio (PSNR) and Structural Similarity (SSIM) are used to estimate the performance in our experiments. 

\subsection{Results on Synthetic Data}
To evaluate the performance of proposed HATNet,
we compare it with previous methods, including ReconNet~\cite{kulkarni2016reconnet}, ISTA-Net$^+$~\cite{zhang2018ista}, CSNet$^+$~\cite{shi2019image}, SCSNet~\cite{shi2019scalable}, OPINENet$^+$~\cite{zhang2020optimization}, AMP-Net~\cite{zhang2020amp}, TransCS~\cite{shen2022transcs}, MADUN~\cite{song2021memory}, DGUNet$^+$~\cite{mou2022deep}, OCTUF$^+$~\cite{song2023optimization}, and SAUNet~\cite{wang2023saunet}.
\cref{tab: result} reports the average PSNR/SSIM of different methods. Our method outperforms previous methods at all SRs. 
\cref{fig: SR} visualizes the reconstruction results of our HATNet and previous competitive methods. Obviously, our HATNet has a significant improvement in image details and textures, as highlighted in the zoom-in regions.


\subsection{Results on Real Data}


\noindent{\bf SPI Prototype Details.} 
To evaluate the real performance of our proposed method, we build a SPI Prototype to capture real data as illustrated in \cref{fig: setup} (a), which mainly consists of a digital micro-mirror device (DMD), and a single-pixel detector (SPD).
A DMD is used to spatially filter light by selectively redirecting parts of an incident light beam.
A DMD is used to measure the total filtered intensity.
An object is illuminated and imaged onto the DMD, where a sequence of binary patterns displayed on the DMD are used to mask the image, and then integrated into one pixel detected by SPD.
In view of practicality, we use cake-cutting Hadamard matrix (CCH)~\cite{cch, hadamardorder, 2022demosaicing}, a variant of Hadamard matrix, as the measurement matrices, whose each row is a binary pattern to be displayed on the DMD. 

\begin{figure}[t]
  \centering
  \vspace{-2mm}
  \includegraphics[width=1\linewidth]{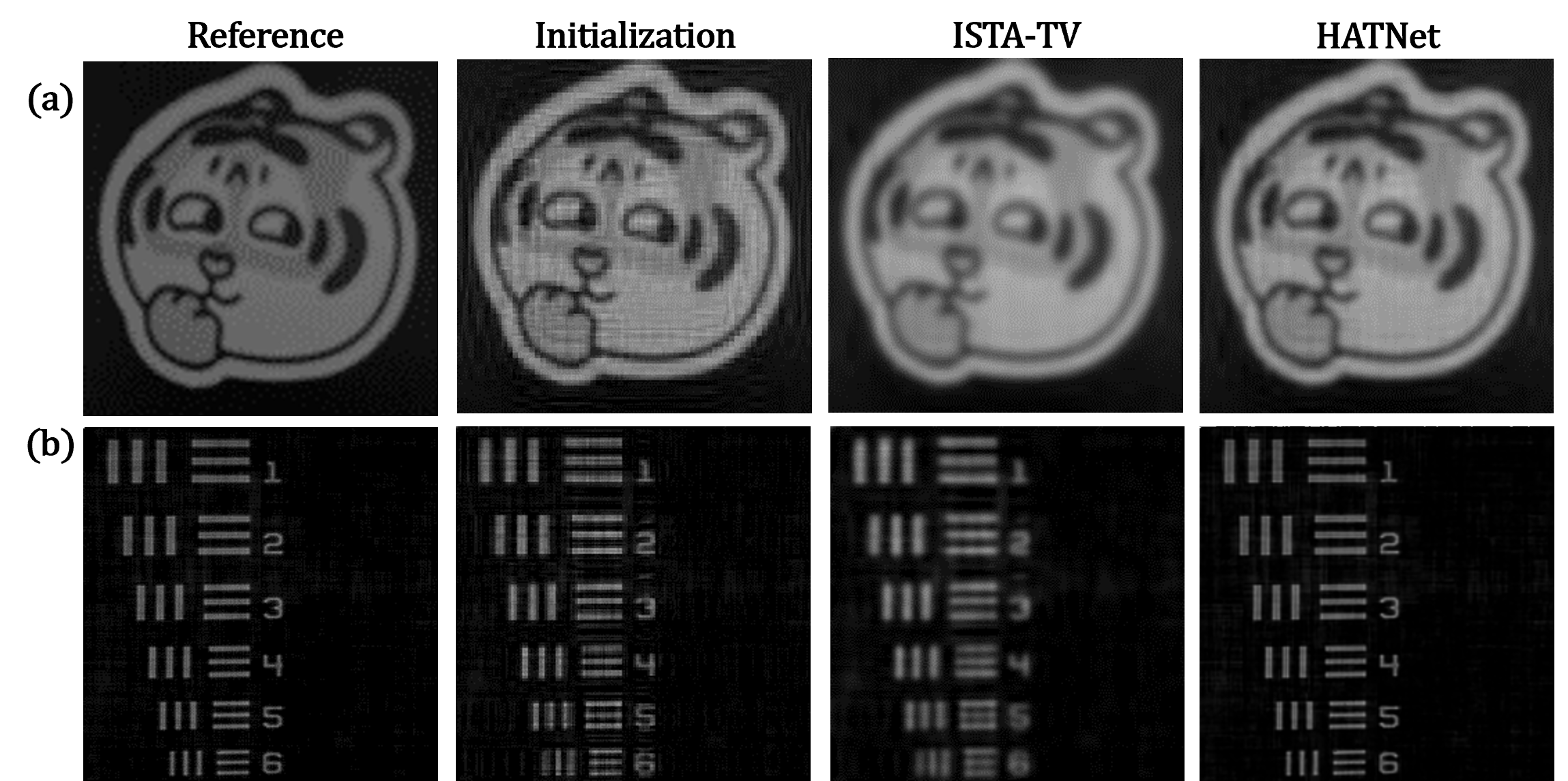}
  \vspace{-5mm}
  \caption{Experimental results of (a) \texttt{cartoon tiger} and (b) \texttt{resolution target} reconstructed by different methods at SR~$=25\%$.}
\label{fig: ER2}
\vspace{-7mm}
\end{figure}

\noindent{\bf Middle-Scale Results.}
We use our SPI prototype to capture real measurements of different scenes with $256\times256$ pixels, and then they are reconstructed by ISTA-TV~\cite{GAPTV}, DGUNet$^+$~\cite{mou2022deep}, OCTUF$^+$~\cite{song2023optimization}, and SAUNet~\cite{wang2023saunet}.
ISTA-TV is a representative optimization algorithm and SAUNet is the first practical deep unfolding network.
Note that original DGUNet$^+$ and OCTUF$^+$ are impractical for real SPI cameras due to their block-based sampling, thus we re-trained them under Kronecker SPI for full-size sampling.
Middle-scale reconstructed results are visualized in \cref{fig: setup} (c) and \cref{fig: ER2}.
The reference images in the first column are captured through full-sampling (\ie, uncompressed) SPI.
Full-sampling SPI can be formulated as $\xv = \Amat^\top\yv$ s.t. $\yv = \Amat\xv$, where $\Amat \!\in\!\mathbbm{R}^{N\times N}$ is a orthogonal Hadamard matrix.
In theory, full-sampled image is lossless due to the orthogonality of Hadamard matrix.
The initialization images are simply computed through $\xv = \Amat^\top\yv$, where $\Amat \!\in\!\mathbbm{R}^{M\times N}$ ($M \!\ll\! N $) is a sub-sampling matrix.
However, noise is not avoidable in real optical system and thus they serve as the references.
Clearly, our method leads to the best visual results.

\begin{figure}[t]
\vspace{-2mm}
\begin{center}
\includegraphics[width=1.0\linewidth]{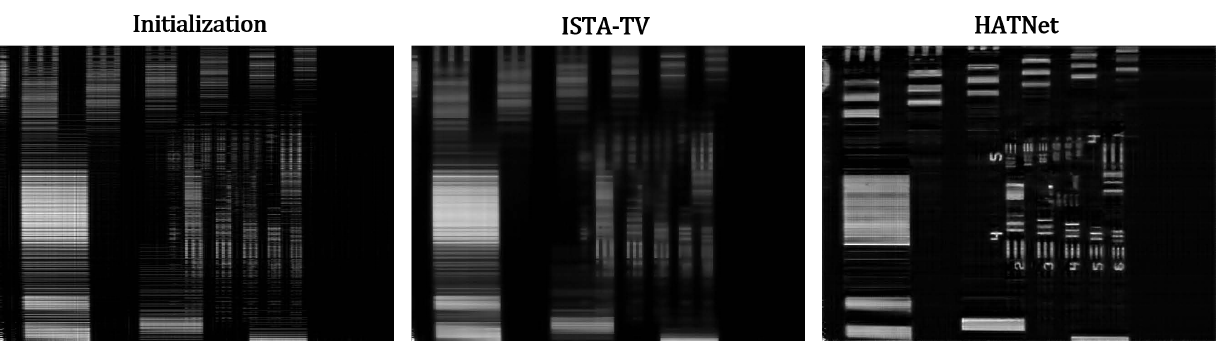}
\end{center}
\vspace{-5mm}
\caption{Large-scale experimental results with $768 \!\times\! 1024$ pixels at SR = $12.5\%$.}
\label{fig:large}
\end{figure}

\vspace{1mm}
\noindent{\bf Large-Scale Results.}
As mentioned previously, previous SPI methods~\cite{wang2022single, qu2022demosaicing, kulkarni2016reconnet}, vectorizing 2D image into 1D signal to process, leads to a huge measurement matrix and thus high computational costs in the forward model and the gradient descent projection, which makes it difficult to train on large-scale images.
Our method utilizes Kronecker SPI to replace a huge measurement matrix with two small measurement matrices.   
The maximal resolution of our DMD is $768 \!\times\! 1024$.
We try to capture image with $768 \!\times\! 1024$ pixels at the sampling ratio $12.5\%$, namely compress 768,432 pixels into 78,304 measurements.
The size of two measurement matrices are $256 \!\times\! 768$ and $384 \!\times\! 1024$.
We use the two matrices to train our HATNet on  20,000 images with $768 \!\times\! 1024$ pixels, which are cropped from DIV2K dataset.
To relieve memory and computational overheads, we properly reduce the number of stage and channel. 
Due to the high training costs, we do not re-train previous methods for comparison.
\cref{fig:large} reports the large-scale reconstructed results of our HATNet and ISTA-TV.
Our HATNet outperforms ISTA-TV by a large margin in the case of large-scale SPI reconstruction.

\begin{figure}[t]
  \centering
  \vspace{-2mm}
  \includegraphics[width=1\linewidth]{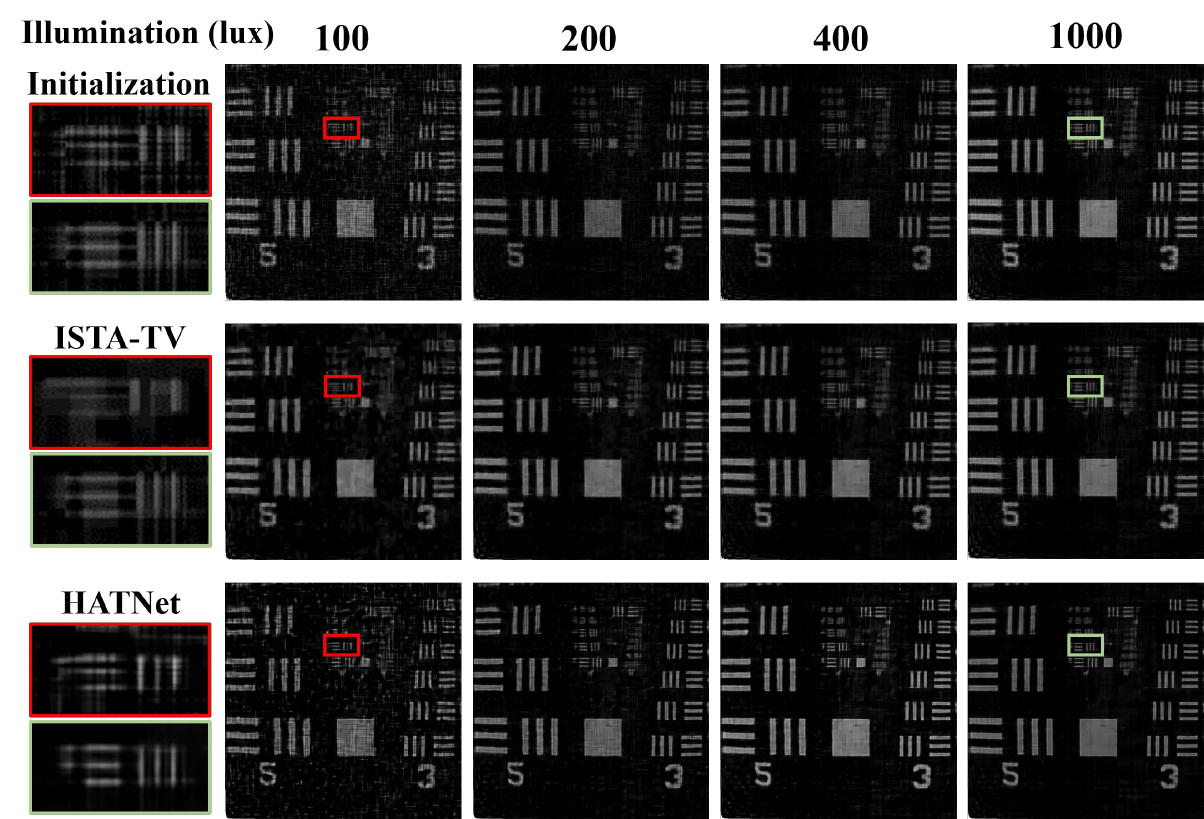}
  \vspace{-5mm}
  \caption{Experimental results of different illumination intensity at SR = $25\%$.}
\label{fig: lx}
\vspace{-3mm}
\end{figure}

\vspace{1.7mm}
\noindent{\bf Illumination-Varying Results.}
We also conduct experiments with varying light intensity, from 100 lux to 1,000 lux, to evaluate the generalization ability of our HATNet.
In general, stronger the illumination intensity is, the higher signal-to-noise-ratio (SNR) is.
\cref{fig: lx} reports the reconstructed results of ISTA-TV and our HATNet in different illumination intensities.
Clearly, the reconstructed results become better as the illumination intensity increases. 
Our HATNet shows a great generalization ability in both low- and high-light conditions. 

\begin{figure}[t]
  \centering
  \includegraphics[width=1\linewidth]{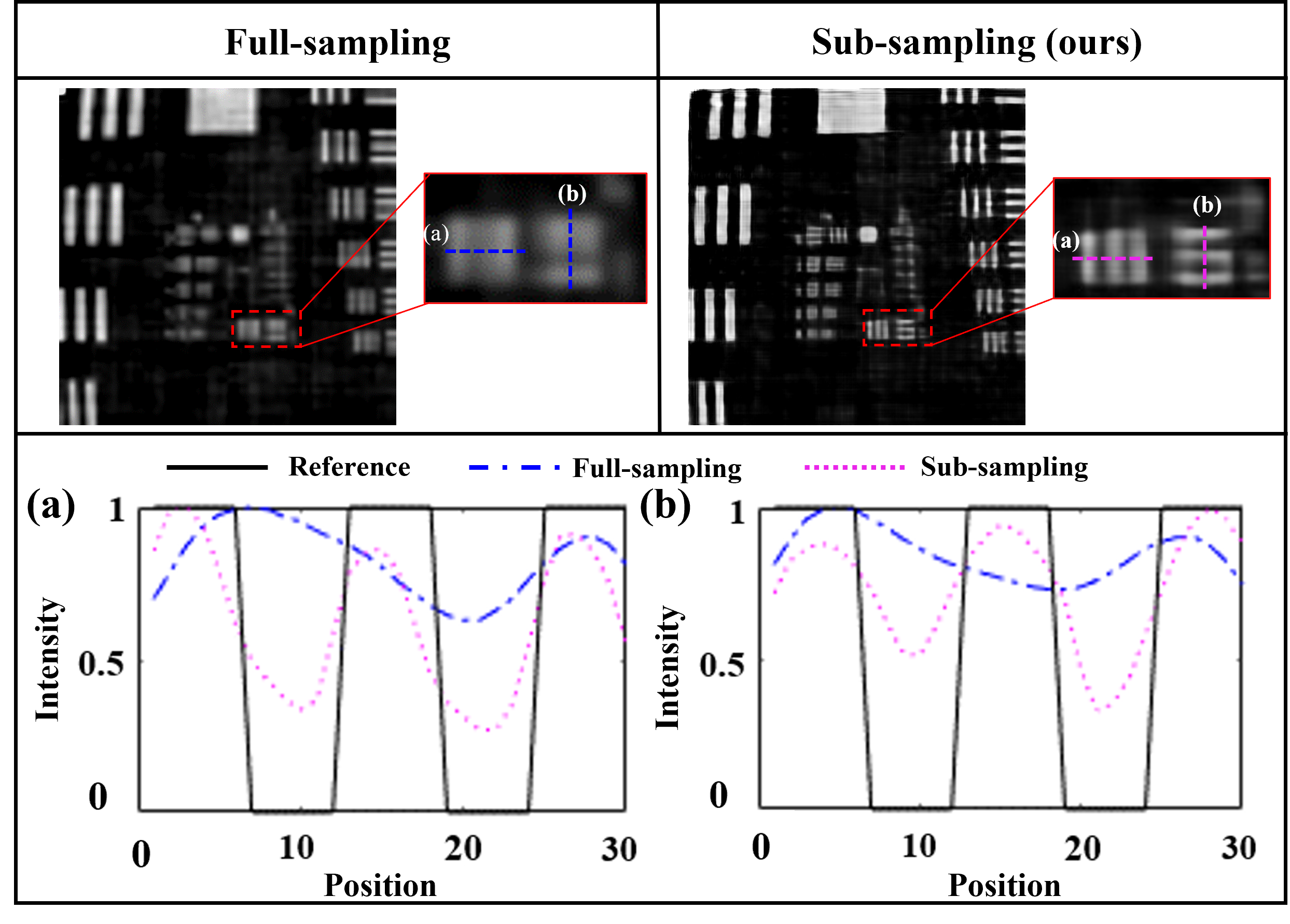}
  \vspace{-6mm}
  \caption{Optical resolution comparison under the same measurements. Full-sampling result (left) versus our sub-sampling result (right). The bottom tables visualize the intensity curve of highlighted lines.}
\label{fig: res}
\vspace{-6mm}
\end{figure}

 
\vspace{1mm}
\noindent{\bf Optical Resolution.}
Under the same data throughput, we are curious whether our sub-sampling method has truly improved the optical resolution compared to full-sampling method.
To this end, we apply full-sampling SPI to capture $4,096$ measurements using a orthogonal Hadamard matrix and then form a theoretically lossless $64\times64$ image.
We apply sub-sampling SPI to capture $4,096$ measurements at SR = $6.25\%$ and then the proposed HATNet to reconstruct a $256\times256$ image.
The full-sampling and sub-sampling images are visualized in \cref{fig: res}.
The results demonstrate our method, a pipeline of sub-sampling plus deep reconstruction, can improve the optical resolution significantly.


%

\vspace{-1mm}

\vspace{1mm}
\subsection{Ablation Study}
\noindent{\bf Different Components of HATNet.}
Our HATNet is mainly powered by the following designs: cross-stage skip connections (CSSC), high-frequency (HF) and low-frequency (LF) aggregation of spatial-wise self-attention (S-SA), and channel-wise self-attention (C-SA).
To demystify the influence of different components, we conduct thorough ablation experiments on Set11 dataset at SR = $10\%$.
The average PSNR and SSIM are reported in \cref{tab: ablation}. Baseline model (a) involves complete components and yields the best result of 32.26 dB/0.9182. Towards model (b) without CSSC, there is an
average 0.78 dB /0.0129 performance degradation, revealing that regular deep unfolding has an inherent loss information issue due to its frequent signal-to-feature transformation.
Towards model (c) with only LF aggregation in S-SA, there is an
average 0.70 dB/0.0108 performance degradation.
Towards model (d) with only HF aggregation in S-SA, there is an
average 0.48 dB/0.0070 performance degradation.
It reveals that the proposed S-SA has a good modeling ability for high- and low-frequency information.
Towards model (c) without C-SA, there is an
average 0.11 dB/0.0042 performance degradation, meaning that our used channel-wise global information recalibration is effective for SPI reconstruction.

\noindent{\bf Kronecker SPI.}
By virtue of Kronecker SPI, our HATNet has advantages over previous vectorized methods~\cite{wang2022single, qu2022demosaicing, kulkarni2016reconnet} and block-based methods~\cite{mou2022deep,song2023optimization}.
Vectorized methods, modeling 2D image into 1D signal, can be deployed in real SPI cameras but their performance is greatly limited by high computational complexity.
For example, ReconNet~\cite{kulkarni2016reconnet} has an average 24.38 dB/0.7301 result at SR = $10\%$, far lower than 32.26 dB/0.9182 of our HATNet.
Block-based methods~\cite{mou2022deep,song2023optimization}, divide image into small-size patches to process and their block-based sampling is impractical in mainstream SPI cameras.
Our HATNet has both practicality and SOTA performance.
We conduct experiments to reveal the superiority of Kronecker SPI.
The comparison between Kronecker SPI and vectorized SPI is shown in \cref{tab: kron}.  
By shifting HATNet from Kronecker SPI to vectorized SPI, GPU memory occupation and inference time increase from 3.02 G and 0.38 s to 10.74 G and 0.55 s, revealing the efficiency of our HATNet.
Full-size sampling of Kronecker SPI is compared with previous block-based sampling as shown in \cref{tab: kronecker}, where we re-train DGUNet$^+$~\cite{mou2022deep} and OCTUF$^+$~\cite{song2023optimization} on Kronecker SPI and re-train HATNet under block-based pipeline.
Clearly, the re-train DGUNet$^+$ and OCTUF$^+$  achieve a clear improvement and the re-train HATNet have a drop on performance, revealing the effectiveness of Kronecker SPI.


\vspace{-3mm}
\begin{table}[H]
\centering
\caption{Ablation study for different components in HATNet.} 
\vspace{-3mm}
\label{tab: ablation}
\renewcommand\tabcolsep{6.5pt}
\scalebox{0.85}{
\begin{tabular}{|c|c|c|c|c|c|c|}
\hline
Model &CSSC & HF & LF & C-SA & PSNR (dB) & SSIM   \\ \hline
(a) & \checkmark  & \checkmark  & \checkmark & \checkmark  & {\bf 32.26} & {\bf 0.9182} \\  \hline 
(b) & &  \checkmark   &  \checkmark  &  \checkmark  & 31.48 & 0.9053 \\ \hline
(c) & \checkmark  &      &    \checkmark    & \checkmark  & 31.56 & 0.9074 \\ \hline
(d) & \checkmark  & \checkmark &  & \checkmark & 31.78 & 0.9112 \\ \hline
(e) & \checkmark  & \checkmark  & \checkmark &   & 32.15 & 0.9140 \\  \hline 
\end{tabular}
}\end{table} 
\vspace{-2mm}
\vspace{-1mm}

\begin{table}[H]
\centering
\vspace{-3mm}
\caption{Comparison between Kronecker SPI and Vectorized SPI at SR$=25\%$.} 
\vspace{-3mm}
\label{tab: kron}
\scalebox{0.62}{
\begin{tabular}{|c|c|c|}
\hline
\multirow{3}{*}{Method} & Kronecker SPI  & Vectorized SPI  \\  \cline{2-3}
{} & $\Xmat \!\in\! \mathbbm{R}^{N\!\times\! N}$, $\Ymat \!\in\! \mathbbm{R}^{M\!\times\! M}$  &  $\xv \!=\! \vc(\Xmat) $, $\yv \!=\! \vc(\Ymat)  $ \\
{} &$\Phimat \!\in\! \mathbbm{R}^{M\!\times\! N}$,  $\Psimat \!\in\! \mathbbm{R}^{M\!\times\! N}$, $\alpha  \!=\! {M^2 \mathord{\left/ {\vphantom {M^2 N^2}} \right. \kern-\nulldelimiterspace} N^2}$    &   $\Amat \!\in\! \mathbbm{R}^{M^{2}\!\times\! N^{2}}$, $\alpha  \!=\! {M^2 \mathord{\left/ {\vphantom {M^2 N^2}} \right. \kern-\nulldelimiterspace} N^2}$     
\\ \hline
Measurement  &   $\Ymat = \Phimat \Xmat \Psimat^\top \Rightarrow \yv = \Amat\xv, \Amat= \Psimat \otimes \Phimat$ &  $\yv = \Amat\xv$    \\ \hline
Gradient descent  &  $\Zmat_{k} \!=\! \Xmat_{k\!-\!1} \!+\! \rho {\Phimat^\top(\Ymat \!-\! \Phimat \Xmat_{k\!-\!1} \Psimat^\top)\Psimat}$  & $\zv_{k} \!=\!  \xv_{k\!-\!1} \!+\! \rho {\Amat^\top(\yv \!-\! \Amat \xv_{k\!-\!1})}$     \\ \hline
Complexity  &  $\mathcal{O}(({\sqrt \alpha } +\alpha )N^3)$    & $\mathcal{O}(\alpha  N^4)$     \\ \hline
GPU memory (G)  &   $3.02$   &   $10.74$  \\ \hline
Inference time (s)  &  $0.38$    &   $0.55$   \\ \hline
\end{tabular}
}
\vspace{-4mm}
\end{table}

\begin{table}[H]
\centering
\vspace{-2mm}
\caption{Comparison between block-based sampling and full-size sampling of Kronecker SPI at SR$=\!10\%$.} 
\vspace{-3mm}
\label{tab: kronecker}
\renewcommand\tabcolsep{6.3pt}
\scalebox{0.8}{
\begin{tabular}{|c|c|c|c|c|c|c|}
\hline
Method & DGUNet $^{+}$ \cite{mou2022deep} & OCTFU$^{+}$ \cite{song2023optimization} & HATNet (ours)  \\ \hline
Block-based & 30.92/0.9088  & 30.73/0.9037    & 31.62/0.9115 \\ \hline
Full-size & 31.65/0.9110 & 31.51/0.9102    & 32.26/0.9154  \\ \hline
\end{tabular}
}
\end{table}

\section{Conclusion}
%


Towards real-world SPI cameras, previous vectorized methods are limited in resolution and performance, and previous block-based methods are impractical. In this paper, we propose a deep unfolding network with hybrid-attention
Transformer on Kronecker SPI model, dubbed HATNet, to realize practicality and SOTA performance.
By unrolling the computation graph of tensor ISTA, HATNet addresses SPI reconstruction problem through two alternative modules: efficient tensor gradient descent and hybrid-attention
Transformer (HAT) based deep denoising.
By virtue of Kronecker SPI, HATNet can efficiently reduce the computational costs, GPU memory, and inference time by replacing a regular large measurement matrix with tow small matrices in the gradient decent projection.
Toward deep denoising module, we propose HAT to aggregate high- and low-frequency information in spatial dimensions and recalibrate global information along channel dimension.
Overall, the proposed method has a potential of improving real-world SPI cameras and take one significant step towards real-world computational imaging applications~\cite{suo2023computational}.

\section*{Acknowledgement}
This work was supported by National Natural Science Foundation of China (62271414), Science Fund for Distinguished Young Scholars of Zhejiang Province (LR23F010001), Research Center for Industries of the Future (RCIF) at Westlake University and and the Key Project of Westlake Institute for Optoelectronics (Grant No. 2023GD007).

{
    \small
    \bibliographystyle{ieeenat_fullname}
    \bibliography{main}
}


\end{document}